# Style Attuned Pre-training and Parameter Efficient Fine-tuning for Spoken Language Understanding

*Jin Cao, Jun Wang, Wael Hamza, Kelly Vanee, Shang-Wen Li*

Amazon AI

`{jincao, juwanga, waelhamz, kvanee, shangwel}@amazon.com`

## Abstract

Neural models have yielded state-of-the-art results in deciphering spoken language understanding (SLU) problems; however, these models require a significant amount of domain-specific labeled examples for training, which is prohibitively expensive. While pre-trained language models like BERT have been shown to capture a massive amount of knowledge by learning from unlabeled corpora and solve SLU using fewer labeled examples for adaption, the encoding of knowledge is implicit and agnostic to downstream tasks. Such encoding results in model inefficiencies in parameter usage: an entirely new model is required for every domain. To address these challenges, we introduce a novel SLU framework, comprising a conversational language modeling (CLM) pre-training task and a light encoder architecture. The CLM pre-training enables networks to capture the representation of the language in conversation style with the presence of ASR errors. The light encoder architecture separates the shared pre-trained networks from the mappings of generally encoded knowledge to specific domains of SLU, allowing for the domain adaptation to be performed solely at the light encoder and thus increasing efficiency. With the framework, we match the performance of state-of-the-art SLU results on Alexa internal datasets and on two public ones (ATIS, SNIPS), adding only 4.4% parameters per task.

## 1. Introduction and related works

Spoken language understanding (SLU), traditionally consisting of intent classification (IC) and slot labeling (SL) tasks, has drawn the attention of researchers as a core component of the goal-oriented dialogue system [1–3]. IC is the task of classifying the utterance of a user into an intent label, such as GetWeather or PlayMusic, while SL is a domain-dependent sequential labeling task. SL aims to label the span of tokens in the utterance for each slot type associated with the intent in the domain ontology. Studies in SLU have led to many successful industrial applications, including Alexa, Google Assistant, Siri, and Cortana [4].

As of late, most state-of-the-art SLU systems are based on deep learning models [5–7]. These neural models provide substantial performance gains, but usually require a large number of labeled utterances for training. The requirement hinders the application of neural SLU systems to different domains, as collecting in-domain data is expensive and time-consuming. Recent advancements in language model pre-training, such as BERT, GPT2, and T5 [8–12], can address the challenges of models being data-hungry. These pre-trained language models learn from massive text corpora in general domains without labeling and store a substantial amount of learned knowledge within parameters. Fine-tuning is then adopted to adapt encoded general knowledge to specific downstream tasks while using significantly fewer labeled examples. This pre-training-fine-tuning technique has yielded significant improvements in resolving SLU and many other natural language understanding problems, such as translation and question-and-answer tasks.

Although these pre-trained models are powerful, they require large amounts of parameters and space to store the knowledge, since semantic and syntactic information is implicitly learned from the co-occurrence of words in utterances and segments [13]. To improve performance in completing downstream tasks, one has to train ever-larger models with additional data, in hopes of capturing more knowledge that overlaps with desired tasks. The resulting process is prohibitively expensive and slow for the research and deployment of SLU systems at both the training and inference stages. Furthermore, the application of SLU generally consists of many downstream tasks, where each task solves for IC and SL within a given domain. Fine-tuning is parameter-inefficient in such applications, as the entire model has to be updated for each task.

To model knowledge more efficiently, we propose a novel network framework that achieves better performance than existing frameworks on SLU downstream tasks and can be extended to solve for new domains in a parameter-efficient manner. This framework consists of a conversational language modeling (CLM) pre-training task and a light encoder architecture. CLM allows the model to be pre-trained using tasks with styles attuned to downstream tasks; the light encoder that aggregates and encodes knowledge represented within the pre-trained language model (in this paper, a CLM style BERT) at various levels of granularity.

Style attuned pre-training consists of pre-training models using focused data and tasks matched for a given set of downstream problems, resulting in models learning more relevant knowledge for those downstream problems. Style attuned pre-training has shown improved outcomes in solving many human language technology problems. Retrieval-augmented language models are directly optimized to retrieve and attend over documents during pre-training and, after fine-tuning, achieves better performance on open-domain question-and-answer tasks [13]. Translation language modeling pre-trains networks with pairs of parallel sentences in different languages to facilitate cross-lingual alignment learning in models and thus outperforms prior machine translation approaches [14]. Here, we propose using CLM to pre-train models with pairs of user queries and system responses, such that networks can encode the representation of language in conversation styles. Furthermore, queries decoded by automatic speech recognition (ASR) systems are used for models to learn ASR error robust knowledge.

In addition to CLM pre-training, we utilize a light encoder to transform generally encoded knowledge into domain- and task-dependent knowledge in a more parameter efficient manner. There are numerous research efforts on improving the ef-ficiency of the pre-trained models. TinyBERT [15] proposed a knowledge distillation framework by training a smaller network to match the output from different layers of a larger frozen model and showed that more than 96% of the performance of the larger model can be recovered by the distilled model. ELMo [16] proposed to learn a linear weighted combination of the layers that would be used as inputs to a task-specific model. Adapter modules [17] also have been proposed for a compact and extensible model, where each adapter consists of a few trainable parameters added in every layer of the frozen pre-trained BERT and is dedicated to one specific task. In this manner, new tasks can be added to these models without interfering with existing ones. Here, we propose a multi-stage style attuned pre-training and parameter efficient fine-tuning setup for downstream tasks. We then conduct an exhaustive set of experiments and demonstrate that the proposed framework outperforms public available approaches, on IC and SL tasks across various domains.

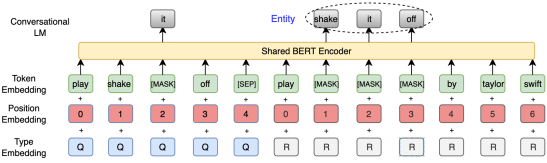

Figure 1: *CLM style BERT pre-training*

In summary, our primary contributions are three-fold:

- Propose a CLM style pre-training task tailored to SLU problems. The task allows models to learn language representation in conversation style with the presence of ASR errors;
- Introduce a novel parameter efficient model architecture comprised of a pre-trained model and a light encoder;
- Perform an empirical comparison among various pre-training and network architectures on model accuracy across IC and SL tasks. These experiments show that our proposed parameter efficient architecture yields comparable accuracy with the SOTA approach, which fine-tunes an entire model per domain.

## 2. Approach

In this section, we describe the proposed network framework, which consists of CLM pre-training and light-encoder fine-tuning. Using the former, we aim to enhance SLU performance via pre-training models with tasks more attuned to downstream problems. The latter enables us to build domain-specific SLU models more efficiently in terms of parameter usage.

### 2.1. CLM Pre-training

Language model pre-training allows a network to learn general language information effectively without human-annotated data. Pre-trained models can then solve for a set of downstream tasks by adapting the models with only a small amount of annotated data. Transformer-based, deep-neural models such as BERT or T5 are widely used to learn and store general language information. In this paper, we choose BERT as the base architecture model, due to a more balanced trade-off between model size and performance as compared with the T5 model. Conventionally, BERT is pre-trained with a masked language model and next sentence prediction tasks using public data sets including Wikipedia. However, networks can only learn general, written-form language information in such pre-training.

In light of this limitation, here, we propose to use a conversational language model, CLM, for networks to encode knowledge attuned to SLU tasks, as shown in Figure 1. To build CLM, we first collected a dataset consisting of conversations between users and a goal-oriented dialog system. An example of a conversation begins with a user query of *"play Shake It Off"*, followed by the system responding with*"play Shake It Off by Taylor Swift"*. In addition to semantic and syntactic information, this dataset provides knowledge about the conversational context distilled from the dialog system components, such as natural language generator (e.g., *"play song by artist"* is a response to the query *"play song"*), entity resolution (e.g., *"Shake It Off"* is a song), and knowledge graph (e.g., *"Taylor Swift"* is the artist of *"Shake It Off"*). We then concatenate each query-response pair into one sequence and append type embedding to each token embedding as showed Figure 1. Type embedding is a binary value to distinguish the query from the response. Lastly, we apply masking to entities in the respective queries and responses. In the above example, *"Shake It Off"* are masked (partially) in both query and response. Entity masking forces models to relate entities using conversational context distilled from system components. With such a setting, the models accumulate knowledge in line with both the dialog system as well as its SLU component. An additional benefit in using this type of conversational dataset is that the user queries in these datasets are decoded with an ASR system that could generate recognition errors, while the system response is free of ASR errors. The model can thus be systematically trained on ASR errors and any learned knowledge is robust against errors.

Pre-training with conversational data also improves the ability of a model to use generalized language information in solving downstream SLU tasks, especially for domains with limited amounts of labeled data. It is well-established that unsupervised pre-training allows models to encode general information, while supervised fine-tuning helps models to learn domain-specific information for downstream tasks. However, a realistic SLU system comprises many domains and, inevitably, some domains have fewer annotated labels than others. Imbalanced supervised signals result in biased performance across the domains. By pre-training with conversations between users and dialog system, models can learn language information shared amongst domains within a given system. Domains with limited annotation can leverage information from similar domains within the SLU/dialog realm, improving the generalizability of networks.

### 2.2. Light Encoder Fine-tuning

A popular approach to adapt a pre-trained model to downstream tasks is fine-tuning with domain-specific data. However, building models for each domain is prohibitively expensive, especially for models leveraging large networks such as BERT. Thus, to build parameter-efficient models we further propose an architecture consists of a sharable, pre-trained network and a light, task-dependent encoder. As an SLU system grows, the resulting models are scalable to the increasing number of domains and maintain stable performance throughout the expansion.

The architecture of the proposed light encoder and its application to the pre-trained network is presented in Figure 2. The blue blocks on the left stand for the frozen transformer layers that are pooled together and shared across domains. The green blocks on the right illustrate the light encoder, which includes the layers tuned for downstream tasks (i.e., for each SLU model in one domain). As shown in the figure, we first transform utterances into contextual token-level vectors with BERT and then concatenate vectors from each of the BERT internal layers to obtain the token-level, domain-agnostic representation. Since parameters are frozen in this stage, we can share the pooled representations with different downstream tasks. The representation is subsequently transformed with a dense layer and a BiLSTM block. We further feed the utterance representation to the output layers, consisting of a softmax layer for IC and a Conditional Random Fields (CRF) layer for SL, for prediction. When fine-tuning for downstream tasks, only the parameters in the light encoder (i.e., the green cells) are trained. Thus, for each domain, we only need to maintain parameters for the light encoder. The parameters for BERT are stored once and shared across all domains. We refer to this architecture as Concat+LSTM in this paper. We propose to concatenate vectors from each of the BERT layers as prior studies, ELMo [16] and BERT adaptor [17], have shown that utilizing vectors from different encoder layers results in improved model performance, as different layers may capture a variety of linguistic information. Lower layers are likely to capture syntactic clues, while higher layers encompass semantic information.

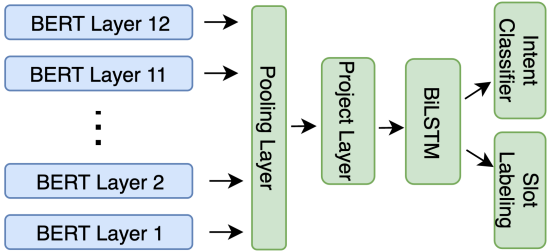

Figure 2: *Overview of proposed light encoder architecture*

To understand the performance impact of each component in the proposed model, we investigated three additional architectures: Linear+LSTM, LastLayer+LSTM, and Concat. The Linear+LSTM approach uses a learned linear weight vector to combine representation in each internal layer of BERT before the dense projection layer and BiLSTM block. The LastLayer+LSTM architecture utilizes the representation from the last layer of BERT directly. As for the Concat architecture, we remove the BiLSTM block from our proposed Concat+BiLSTM architecture. All parameters in the light encoder, except the ones in the output layers, are initialized by training the encoder with the frozen BERT using the masked language model task.

Table 1: *Fine-tuning Dataset*

| Dataset | #train-size | #test-size | #intents | #slots |
|---|---|---|---|---|
| ATIS | 4.1k | 0.8k | 16 | 65 |
| SNIPS | 11.8k | 0.6k | 7 | 47 |
| Music | 100k | 72k | 22 | 79 |
| Shopping | 100k | 72k | 16 | 27 |
| Localsearch | 100k | 12k | 22 | 47 |

## 3. Experiments

### 3.1. Dataset

We evaluated the performance of proposed approaches with IC and SL tasks using two public datasets (SNIPS [18] and the Airline Travel Information System corpus (ATIS) [19]) and three Alexa internal datasets (Local-Search, Music, and Shopping). SNIPS consists of utterances with seven intents across multiple domains, including music, media, and weather, and was used to evaluate the accuracy of NLU services. ATIS contains transcribed audio recordings of individuals making flight reservations and spans across seventeen intent categories, such as flight booking or aircraft capacity inquiries. For internal datasets, we sampled Alexa daily traffic from music, shopping, and local search domains. Utterances in these Alexa internal datasets consist of various entities, such as artists, albums, cities, locations, business types, and product attributes. We list detailed statistics (e.g., number of utterances in training and test sets and number of intents and slots) for the five datasets in Table 1.

Utterances in Alexa internal datasets are user queries transcribed by Alexa ASR system and the IC and SL labels were annotated to the transcribed dialog on hypotheses. In contrast, both public datasets only provide IC and SL labels using human transcription, and SNIPS only offers user queries in text. To evaluate our network architecture on SLU performance, we prepare our own ASR hypotheses as well as IC and SL labels on the underlying data for the two public datasets following a popular data simulation [20]. We first synthesized audio for text queries in SNIPS with a commercial TTS service[1]. We then decoded audio with a commercial ASR system[2] to obtain ASR hypotheses of user queries for both SNIPS and ATIS. The word error rate (WER) of decoded audio for ATIS and SNIPS is 18.4% and 16.2% respectively. We used IC annotation on transcription directly, as the annotation on hypotheses and SL annotation on transcription was projected with Levenshtein alignment for labels on hypotheses.

### 3.2. Experiment setup

We benchmarked our model performances on the above datasets with a public, well-trained BERT model from Google[3] as the baseline (denoted as BERT-Google). This model adopts the standard BERT architecture, with 12 transformer layers, 768 hidden units, and 12 attention heads, and was trained on public data, including Wikipedia, Bookcorpus, and OpenwebText without casing. We also pre-trained BERT with the same model architecture, using a mixture of public corpora and conversational data, on the proposed CLM task (denoted as BERT-CLM). The conversational data was collected from Alexa live traffic covering all the domains and is comprised of 50 million query-response pairs. The user requests within the conversational data were recognized by the Alexa ASR system and tokenized into a spoken format, with no human annotation is required. For an ablation study, a BERT using the same architecture but trained with only the public dataset and user queries in the conversational data was also built (denoted as BERT-Query).

We adopt a 2-layer BiLSTM with 256 hidden units in our light encoder architectures. We use a dense layer of the same hidden size, projecting the output from the pooling layer to match the input of BiLSTM. In the Concat+LSTM architecture, we concatenate the 12 transformer layers output, while, in the Linear+LSTM architecture, we use a linear vector to combine the 12 transformer layers from BERT. We jointly train the downstream IC and SL tasks, using a softmax layer as the IC output layer, and a CRF layer for SL output.

---
[1]Amazon Polly, https://aws.amazon.com/polly/
[2]Amazon Transcribe, https://aws.amazon.com/transcribe/
[3]Google BERT, https://github.com/google-research/bert

Table 2: *Results of applying various pre-trained models on internal and public IC and SL benchmark datasets. Baseline numbers are fine-tuned on BERT-google. Numbers for the internal dataset are relative F1 score differences in percentage as compared to the baseline, and a positive number indicates improvements from baseline. The numbers in ATIS and SNIPS are f1 scores.*

| Model | LocalSearch rel IC/SL | Music rel IC/SL | Shopping rel IC/SL | ATIS IC/SL | SNIPS IC/SL | Avg rel IC/SL |
|---|---|---|---|---|---|---|
| BERT-Google | - | - | - | 96.79 / 93.25 | 98.45 / 88.66 | |
| BERT-Query | -0.50 / 4.10 | 0.50 / 2.40 | 0.80 / 1.90 | 97.02 / 93.43 | **98.61** / 88.86 | 0.2 / 1.8 |
| BERT-CLM | **0.10 / 5.30** | **0.80 / 2.50** | **1.30 / 3.50** | **97.14 / 93.84** | 98.60 / **89.34** | **0.6 / 2.5** |

Table 3: *Results of applying proposed light encoder architecture on internal and public IC and SL benchmark datasets. Baseline numbers are the fine-tuning on BERT-CLM, which is pre-trained with query and response pair. Numbers for the internal dataset are relative F1 score differences in percentage as compared to baseline, while the numbers for ATIS and SNIPS are f1 score.*

| Model | Total num params | Trained params per task | LocalSearch rel IC/SL | Music rel IC/SL | Shopping rel IC/SL | ATIS IC/SL | SNIPS IC/SL | Avg rel IC/SL |
|---|---|---|---|---|---|---|---|---|
| BERT-CLM | 5.0x | 100.0% | - | - | - | 97.14 / 93.84 | 98.60 / **89.34** | - |
| Concat+LSTM | 1.22x | 4.4% | 2.1 / -1.1 | 0.0 / **1.6** | -0.8 / -0.8 | 97.02 / 93.82 | 98.76 / 89.26 | **0.3** / -0.1 |
| Linear+LSTM | 1.13x | 2.5% | **2.2** / -1.2 | 0.0 / 0.8 | -0.8 / -1.1 | 97.12 / 93.29 | 98.76 / 88.69 | **0.3** / -0.6 |
| LastLayer+LSTM | 1.13x | 2.5% | -0.2 / -0.9 | -0.3 / 1.4 | -0.6 / -1.6 | 96.31 / 93 | **99.07** / 86.69 | -0.3 / -1.0 |
| Concat | 1.10x | 2.0% | 1.2 / -3.6 | **0.2** / 0.6 | -0.5 / -2.1 | 95.83 / **90.15** | 98.76 / 86.35 | -0.1 / -2.5 |

### 3.3. Results and discussion

In this section, we present and analyze the experiment results on both the public (ATIS and SNIPS) and internal (Local-Search, Music, and Shopping) datasets. We report F1 scores for each of the datasets on IC and SL tasks separately, to evaluate the benefit of style attuned pre-training and light encoder architectures. We also examine the contributions to the accuracy performance of the different components in the proposed architecture.

Table 2 shows IC/SL performance for BERT models pre-trained with various techniques and fine-tuned to different domains. We observe improved performance in both IC and SL tasks across all domains on fine-tuned BERT-CLM, as compared to the baseline model BERT-Google. The BERT-Query model also outperforms the baseline, although it yields lower F1 scores than the BERT-CLM model. Although our pre-training corpus did not contain utterances from ATIS or SNIPS data, we see improvements for the two domains. We believe our CLM setup facilitates the model to learn the relationship between the entities in query and response, thus allowing the model to learn a representation better fits SLU tasks.

Next, we investigate various light encoder architectures on top of the best-performing pre-trained model (BERT-CLM). As reported in Table 3, our Concat+LSTM architecture achieved comparable performance with fine-tuning the whole BERT-CLM model (an average of 0.3% relative improvement on IC, 0.1% relative degradation on SL) by only using 4.4% of incremental parameters per domain. As for the Linear+LSTM architecture, we observe a slight degradation in results (average of 0.3% relative improvement on IC, 0.6% relative degradation on SL) compared with the baseline, BERT-CLM. We believe the concatenation approach allows the model to retain both the syntactic information from lower layers and the semantic information from higher layers, whilst the linear weight combination approach loses partial information and results in a slight degradation in performance. However, the Linear+LSTM architecture only utilized 2.5% of extra parameters per domain. We also remove the concatenation pooling layer and directly use the final output layer from BERT with LSTM (LastLayer+LSTM), we note a performance degradation on both tasks (average of 0.3% relative degradation on IC, 1.0% relative degradation on SL). We presume that lower layers of BERT capture syntactic information of the utterance, which is especially helpful for SL tasks. Lastly, we remove the BiLSTM encoder (Concat), and we again observe the performance on SL tasks degrades by 2.5%, indicating that the addition of the BiLSTM block allows the model to better utilize the signals captured by the shared BERT encoder and thus benefits the downstream SL tasks.

## 4. Conclusions

With the rising popularity of transfer learning in SLU, the challenge of how to adapt pre-trained models to SLU tasks effectively and efficiently is increasingly relevant. To address the challenge, we propose a novel framework comprised of a style attuned pre-training and light encoder fine-tuning architecture. The style attuned pre-training facilitates model learning of related knowledge for downstream tasks, while the light encoder architecture enables parameter sharing and efficient fine-tuning for the tasks. We demonstrate that BERT pre-trained with conversational style corpora outperforms the publicly available BERT in SLU tasks after fine-tuning the entire networks. The light encoder architecture achieves comparable performance to our best fully fine-tuned model while utilizing less than 5% of incremental parameters per domain. This framework enables domain expansion for SLU at a much lower cost.